\documentclass{article}

\usepackage{PRIMEarxiv}

\usepackage[utf8]{inputenc} 
\usepackage[T1]{fontenc}    
\usepackage{hyperref}       
\usepackage{url}            
\usepackage{booktabs}       
\usepackage{amsfonts}       
\usepackage{nicefrac}       
\usepackage{microtype}      
\usepackage{lipsum}
\usepackage{fancyhdr}       
\usepackage{graphicx}       
\graphicspath{{media/}}     
\usepackage{amsmath} 
\usepackage{float} 
\usepackage[numbers,sort&compress]{natbib}

\title{Pavement Missing Condition Data Imputation through Collective Learning-Based Graph Neural Networks}

\author{
  Ke Yu \\
  School of Computing and Information \\
  University of Pittsburgh \\
  \texttt{key44@pitt.edu} \\
   \And
  Lu Gao \\
  Department of Civil \& Environmental Engineering \\
  University of Houston \\
  \texttt{lgao5@central.uh.edu} \\
}

\begin{document}
\maketitle

\begin{abstract}
Pavement condition data is important in providing information regarding the current state of the road network and in determining the needs of maintenance and rehabilitation treatments. However, the condition data is often incomplete due to various reasons such as sensor errors and non-periodic inspection schedules. Missing data, especially data missing systematically, presents loss of information, reduces statistical power, and introduces biased assessment. Existing methods in dealing with missing data usually discard entire data points with missing values or impute through data correlation. In this paper, we used a collective learning-based Graph Convolutional Networks, which integrates both features of adjacent sections and dependencies between observed section conditions to learn missing condition values. Unlike other variants of graph neural networks, the proposed approach is able to capture dependent relationship between the conditions of adjacent pavement sections. In the case study, pavement condition data collected from Texas Department of Transportation Austin District were used. Experiments show that the proposed model was able to produce promising results in imputing the missing data.
\end{abstract}


\section{Introduction}
Accurate assessment and prediction of pavement asset performance is essential in managing the maintenance and rehabilitation treatments for a road network \cite{Li2005ProbabilisticAdaptive,george1989pavementdeterioration,dhatrak2020considering,burr1987markovpci,medury2014sno,kobayashi2012hiddenmarkov,saha2017markovdistress,almansour2022expertsystem,vemuri2020pavement,rejani2021clustering,khichad2025overview,chen2025rnnltpp,gao2015milled,zhang2018nested,gao2012bayesian,shahid2025stochasticreview,gao2019impacts}. However, pavement condition values are often missing due to various reasons such as sensor failure or non-periodic inspection. Missing data can affect the accuracy of predicting future pavement performance if not handled appropriately. Handling missing predictor values has been heavily studied. There are three general approaches to address missing data in pavement management: (1) eliminate data point with missing values \cite{Piryonesi2020DataAnalyticsIAM}; (2) impute missing values with simple interpolation \cite{Karlaftis2015CrackInitiation}; (3) impute missing values using statistical models by taking known values and historical condition data and other factors into consideration \cite{Gao2011MaintenanceIntervention,Saliminejad2012SpatialBayesian,AlZoubi2015Systematic,Farhan2015Improved,Hafez2016Utilizing,Gao2022MissingImputationGNN}.

In this research, we investigated applying a collective learning-based Graph Neural Network (CLGNN) for imputing missing pavement condition data. Graph Neural Network (GNN) is one of the fastest growing areas in deep learning and it’s designed to utilize the graph structure of data in network format. The proposed model belongs to the third approach discussed above where historical data and neighboring sections information are used for missing value imputation. CLGNN was first developed by \cite{Hang2021CollectiveLearning}. Unlike traditional GNN model, where the feature dependencies of adjacent nodes are modeled, CLGCN takes dependencies between node labels into consideration, which is especially suitable for pavement missing condition data imputation. The objective of this research is to study if the imputation of a pavement section’s missing condition values can be improved by taking neighboring sections’ condition information into consideration.

\section{Methodology}

\begin{figure}[H]
    \centering
    \includegraphics[width=0.75\linewidth]{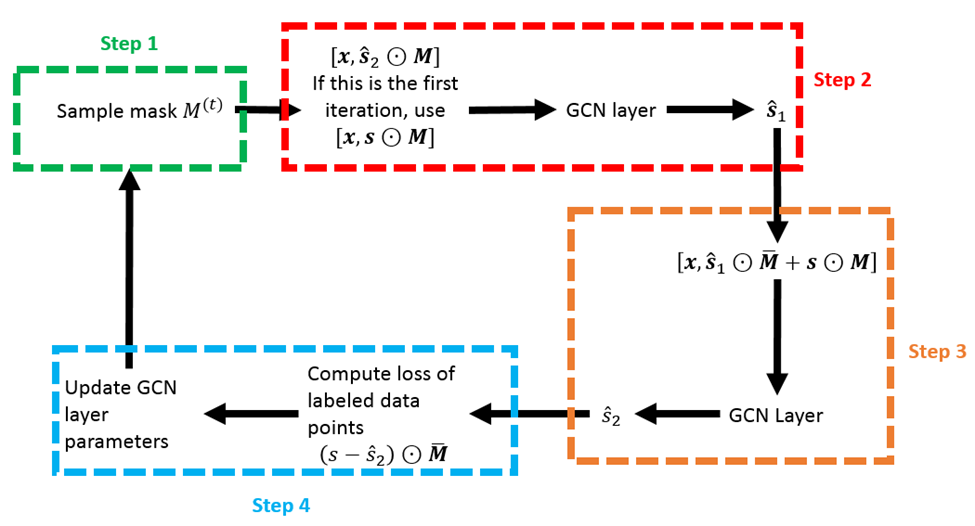}
    \caption{Collective Learning-based GNN framework}
    \label{fig:fig1}
\end{figure}

Applying deep learning models, especially Convolutional Neural Network (CNN) and Long Short-Term Memory (LSTM) models, for pavement performance modeling has received considerable attention recently. Compared with traditional models, deep learning models were found to produce more accurate prediction results
\cite{Lee2019DeepLearningDeteriorationPlanning,Choi2019Deterioration, Hosseini2020DeepLearningDeterioration,Gao2021MaintenanceTreatmentsDetection,Zhou2021RoughnessDL,hosseini2020iowa_lstm,choi2020pavement_dl,justosilva2021ml_review,li2024transfer_pavement,sensors2025modulus_resrnn,hou2025lstm_pqi,yao2019trr_nn_calibration,gong2018deepnn_meepdg_rutting,chen2025cpo_cnn_lstm_dba,sun2024explainable_lstm,hadjattou2023measurement_cnnlstm,zhang2024circle_lstm_rutting,chen2025rnn_ltpp,ahmed2025pinn_iri,deng2023rnn_maintenance_beijing}. Unlike CNN and LSTM where data points are treated as independent objects, GNN is able to utilize the structural information in the network. While GNN models have recently been successfully used for node and graph classification tasks, GNN models dependencies between attributes of adjacent nodes, not dependencies between observed node labels. In this paper, we used a collective learning-based graph neural network model \cite{Hang2021CollectiveLearning}, which consider using GNNs for inductive node classification under supervised and semi-supervised conditions considering label dependencies.We begin by considering a general missing data imputation model. Let $s \in \mathbb{R}^N$ be the condition assessment of a pavement network of $N$ sections.

\begin{equation}
s^{\text{unlabeled}} = f\!\left(s^{\text{labeled}},\, x,\, G\right)
\tag{1}
\end{equation}

where $f$ is the imputation function that predicts the missing (unlabeled) pavement condition.
Here, $s^{\text{unlabeled}}$ and $s^{\text{labeled}}$ are the missing and observed condition vectors, respectively.
The vector $x$ contains explanatory variables such as condition from previous years, traffic, age, and pavement type.
We define the pavement network as a graph $G=(V,E)$ with $N$ nodes $v_i \in V$ representing sections, and edges $(v_i, v_j) \in E$ indicating the connections between sections.
Figure \ref{fig:fig1} shows the overall framework, which consists of the following iterative four steps:
\begin{itemize}
  \item Step 1. Sample a random binary mask;
  \item Step 2. Obtain the predicted label distribution using a Graph Convolutional Network (GCN) layer;
  \item Step 3. Combine the predicted labels with available true labels, and use the result again as input to a GCN layer;
  \item Step 4. Perform parameter optimization by minimizing the loss.
\end{itemize}

At each training iteration, the GCN layer can be expressed as
\begin{equation}
h_i^{(k)} \;=\; \sum_{j \in \mathcal{N}(i)\,\cup\,\{i\}}
\frac{1}{\sqrt{\deg(i)}\,\sqrt{\deg(j)}} \,\Bigl(\Theta\, h_j^{(k-1)}\Bigr)
\tag{2}
\end{equation}
where $h_i^{(k)}$ is the GCN output of node $i$ at layer $k$, $\mathcal{N}(i)$ is the neighborhood of node $i$, $h_j^{(k-1)}$ represents the input embedding of node $j$ (from the previous layer), $\deg(i)$ is the degree of node $i$, and $\Theta$ is the parameter to be estimated.

\section{Case Study}

\subsection{Data Description}
To demonstrate and evaluate the applicability of the proposed model, a case study was carried out using pavement condition inventory data from Texas Department of Transportation (TxDOT)’s Austin District. The pavement condition data was collected between 2014 to 2018. Each pavement section is labeled with a unique reference marker, which was used create the spatial relationships between a section and its neighbors in this research. The variables used in this study contains key attributes of pavement condition observations as shown in Table (1).  Climate is a significant factor in the deterioration of pavement. However, the road sections analyzed in this study are all located in the Austin area. According to a TxDOT report \cite{Xu2021TX2169882}, all the Austin district's pavement sections are situated in the same region with an average annual temperature range of 61.25 to 70 Fahrenheit (16.25 to 21.11 degrees Celsius) and average annual precipitation between 16 and 38 inches (40.64 to 96.52 cm). Thus, the impact of climate on the pavement is not considered in this study.

\begin{table}[H]
\centering
\caption{Variables used in the analysis.}
\label{tab:variables}
\small
\begin{tabular}{p{0.30\linewidth} p{0.66\linewidth}}
\hline
\textbf{Variable} & \textbf{Description} \\
\hline
Pavement condition score &
Overall pavement condition (distress and ride quality), ranging from 1 (worst) to 100 (best). \\

Type of pavement surface &
Pavement surface type, grouped by similarity of characteristics. \\

Functional class &
Texas highway functional classification: groups and sub-groups based on roadway function. \\

Traffic &
Current 18-kip ESAL for the section; values stored in thousands. \\
\hline
\end{tabular}
\end{table}

To conduct this case study, the condition of a pavement was discretized into five different states according to its condition score (CS) as shown in Equation (3). The distribution of the pavement conditions across the road network is shown in FIGURE 2.

\begin{equation}
\text{Condition States} =
\begin{cases}
\text{very good}, & 90 \le \text{CS} \le 100,\\
\text{good},      & 70 \le \text{CS} < 90,\\
\text{fair},      & 50 \le \text{CS} < 70,\\
\text{poor},      & 35 \le \text{CS} < 50,\\
\text{very poor}, & 0 \le \text{CS} < 35.
\end{cases}
\tag{3}
\end{equation}

\begin{figure}[H]
    \centering
    \includegraphics[width=0.75\linewidth]{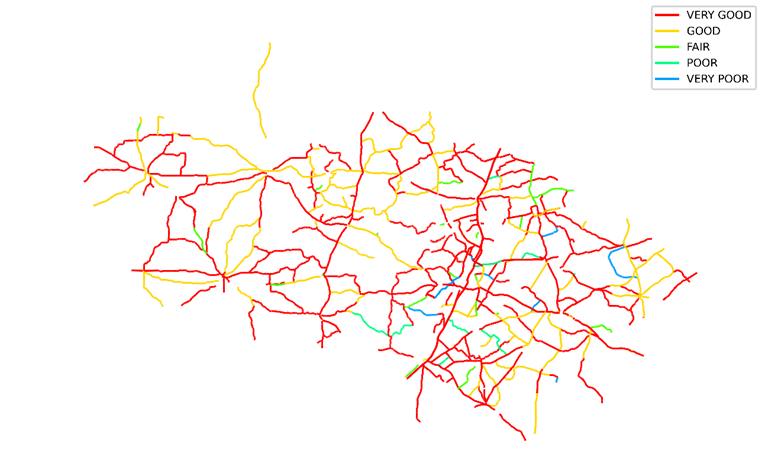}
    \caption{Condition Score Map in the Case Study Database}
    \label{fig:fig2}
\end{figure}

\subsection{Pavement Type}
\begin{figure}[H]
    \centering
    \includegraphics[width=0.75\linewidth]{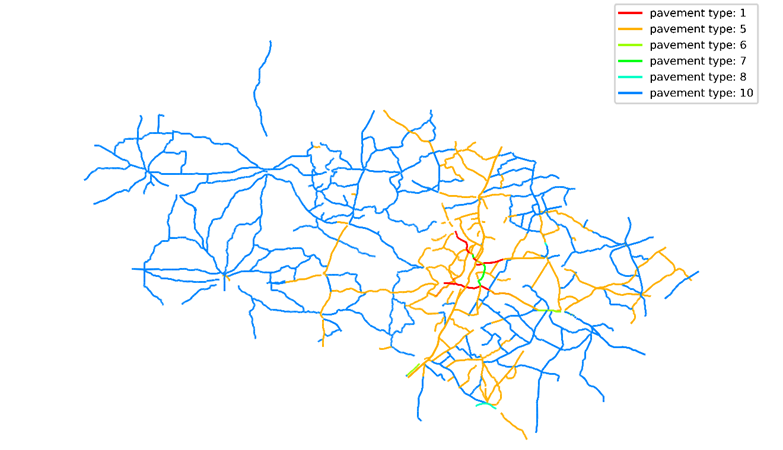}
    \caption{Map of Different Pavement Types }
    \label{fig:fig3}
\end{figure}

There are ten different types of pavements in the TxDOT pavement management system. In this research, six pavement types (codes: 1, 5, 6, 7, 8, and 10) were used (FIGURE \ref{fig:fig3}). Code 1 represents continuously reinforced concrete (CRCP). Code 5 indicates medium thickness asphalt concrete (2.5-5.5”). Code 6 represents thin asphalt concrete (less than 2.5”). Code 7 represents composite (asphalt surfaced concrete on top of heavily stabilized base). Code 8 represents widened composite pavement. Code 10 represents thin surfaced flexible pavement (surface treatment or seal coat). 

\subsection{Functional Classification}
\begin{figure}[H]
    \centering
    \includegraphics[width=0.75\linewidth]{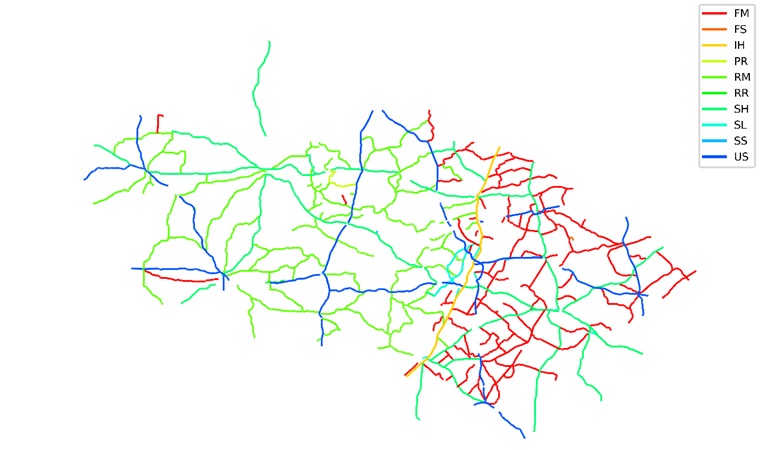}
    \caption{Distribution of Different Functional Classes }
    \label{fig:fig4}
\end{figure}

In Texas, highways are categorized into different groups based on their function. In this case study, 10 different groups of highways are include in the dataset. As shown in FIGURE \ref{fig:fig4}, most of the highways fall into groups of Ranch-to-Market (RM), Farm-to-Market (FM), State Highway (SH), US Highway (US), and Interstate Highway (IH), which corresponds to more than 90 percent of all the records. 

\subsection{Traffic}
In this case study, the 20-year projected ESALs were used to represent the traffic characteristic of each pavement section. The distribution of the traffic in the Austin District was plotted in FIGURE 5.  

\begin{figure}[H]
    \centering
    \includegraphics[width=0.75\linewidth]{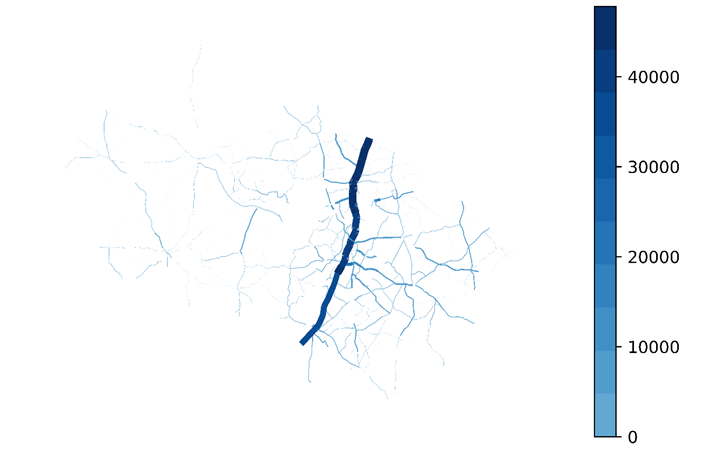}
    \caption{Map of Traffic Distribution}
    \label{fig:fig5}
\end{figure}

\subsection{Models}
\begin{figure}[H]
    \centering
    \includegraphics[width=0.75\linewidth]{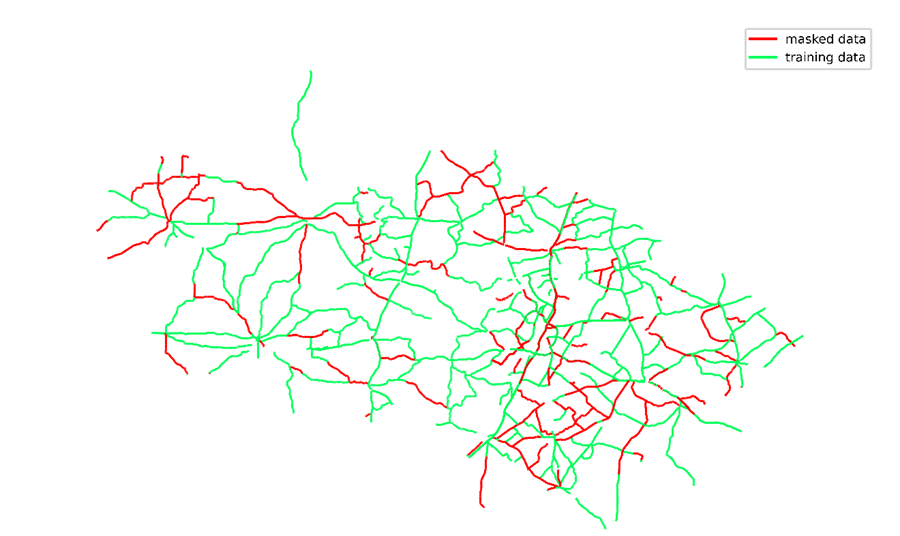}
    \caption{Masked and Training Data}
    \label{fig:fig6}
\end{figure}

In this case study, we developed missing value imputation models for the condition score indicator using the standard machine learning models: classification and regression trees (CART), neural network (NN), and random forest (RF); common GNN models such as Graph Convolutional Network (GCN) and GraphSAGE, and the proposed collective learning based GNN model. All machine learning models were implemented in scikit-learn. GCN, GraphSAGE and CLGNN models were implemented in PyTorch Geometric, which is a Python library supporting many types of deep learning on graphs. We masked 30\% of the 2018 condition scores as missing values and trained the proposed model to predict the masked data points (FIGURE \ref{fig:fig6}). The process of selecting masked sections in the network involves both a random component as well as consideration of the connectivity of the sections. In other words, the sections that are selected to be masked are chosen randomly, but the selection process also takes into account how the sections are connected to each other within the network. The approach of taking section connectivity into consideration when randomly selecting masked sections allows for a more targeted training process. By focusing on clusters of sections within the same route, the network can be trained to better handle missing data in a more realistic and relevant way. This is because in real-world scenarios, missing data is more likely to occur in a cluster of sections within the same route rather than individual sections distributed randomly across the network. By simulating this type of missing data during training, the network can learn to adapt and make more accurate predictions even when faced with missing data in the future. For each road section, the features include 2014-2017 historical condition score data, traffic, road functional class, and pavement type information. 

\subsection{Results}
The modeling results are provided in TABLE \ref{tab:accuracy}, which lists the accuracy of for each model. The proposed CLGNN model, which takes into account both historical condition scores and neighboring sections’ current condition scores, achieves better performance than other models. More specifically, the CLGNN model is able to improve the imputation accuracy by around 5\%.

\begin{table}[H]
\centering
\caption{Classification accuracy comparison across models.}
\label{tab:accuracy}
\small
\begin{tabular}{l c}
\hline
\textbf{Model} & \textbf{Accuracy} \\
\hline
CLGNN     & 0.773 \\
GCN       & 0.725 \\
GraphSAGE & 0.721 \\
RF        & 0.712 \\
CART      & 0.654 \\
NN        & 0.556 \\
\hline
\end{tabular}
\end{table}

\section{CONCLUSIONS}
In this paper, we investigated using collective learning-based graph neural network model to impute pavement missing condition data. The road network is considered as a graph combining historical condition inventory data and spatial connections between neighboring sections. The spatial relationship between a section and its neighboring sections were then taken into account when predicting missing condition scores. The results show that the CLGNN model outperforms other machine learning and deep learning models. Future research is needed to investigate the performance of the proposed model on other missing condition indicators, such as cracking and roughness.

\bibliographystyle{unsrt}  
\bibliography{references}

\end{document}